%% file: arixv_main.tex
\documentclass{article}
\usepackage{spconf,amsmath,graphicx,hyperref}
\usepackage{latexsym}
\usepackage{amssymb}
\usepackage{amsmath}
\usepackage{amsthm}
\usepackage{booktabs}
\usepackage{enumitem}
\usepackage{graphicx}
\usepackage{color}
\usepackage{wrapfig}
\usepackage{float}
\usepackage{amsmath,amssymb,amsfonts}  
\usepackage{hyperref}
\usepackage{algorithm}
\usepackage{algorithmic}
\usepackage{array}
\usepackage[caption=false,font=normalsize,labelfont=sf,textfont=sf]{subfig}
\usepackage{textcomp}
\usepackage{stfloats}
\usepackage{url}
\usepackage{verbatim}
\usepackage{graphicx}
\usepackage{booktabs} 

\title{MAGIA: Sensing Per-Image Signals from Single-Round Averaged Gradients for Label-Inference-Free Gradient Inversion}
\name{Zhanting Zhou, Jinbo Wang, Zeqin Wu and Fengli Zhang}
\address{School of Information and Software Engineering\\
University of Electronic Science and Technology of China\\
Chengdu, Sichuan, China}
\begin{document}
%
\maketitle
\begin{abstract}
We study gradient inversion in the challenging \emph{single-round averaged gradient} (SAG) regime, where per-sample cues are entangled within a single batch-mean gradient. We introduce \textbf{\underline{M}omentum-based \underline{A}daptive correction on \underline{G}radient \underline{I}nversion \underline{A}ttack (MAGIA)}, a novel, label-inference-free framework that \emph{senses} latent per-image signals by probing random data subsets. MAGIA's objective integrates two core innovations: \textbf{(1)} a closed-form combinatorial rescaling that creates a provably tighter optimization bound, and \textbf{(2)} a momentum-based mixing of whole-batch and subset losses to ensure reconstruction robustness. Extensive experiments demonstrate that MAGIA significantly outperforms advance methods, achieving high-fidelity multi-image reconstruction in large-batch scenarios where prior works fail. This is all accomplished with a computational footprint comparable to standard solvers and without requiring any auxiliary information. Co.: \url{https://github.com/itoritsu/MAGIA}.
\end{abstract}
\begin{keywords}
Gradient Inversion Attacks, Federated Learning, Privacy Leakage.
\end{keywords}
\section{Introduction}

\textbf{Background and Threat Model.}
Federated learning (FL) enables on-device training, yet even aggregated updates can leak private data \cite{mcmahan2017fedavg,zhu2019dlg}. We study privacy leakage in the \textbf{single-round averaged-gradient (SAG)} regime \cite{geiping2020inverting,yin2021gradinversion,dimitrov2024spear,carletti2025sok}. Here, an honest-but-curious server observes only the final batch-mean gradient $g^\star$ from an unknown client mini-batch $\mathcal{B}$:
\begin{equation}
\label{eq_agg_grad}
  g^\star \;=\; \tfrac{1}{|\mathcal{B}|}\sum_{(x,y)\in\mathcal{B}}\nabla_\theta \mathcal{L}(\theta;x,y).
\end{equation}
The adversary's goal is to synthesize a dummy batch $\widehat{\mathcal{B}}$ whose gradient matches $g^\star$, thereby reconstructing the private images.

\begin{figure*}[t!]
	\centering
	\includegraphics[width=0.98\linewidth]{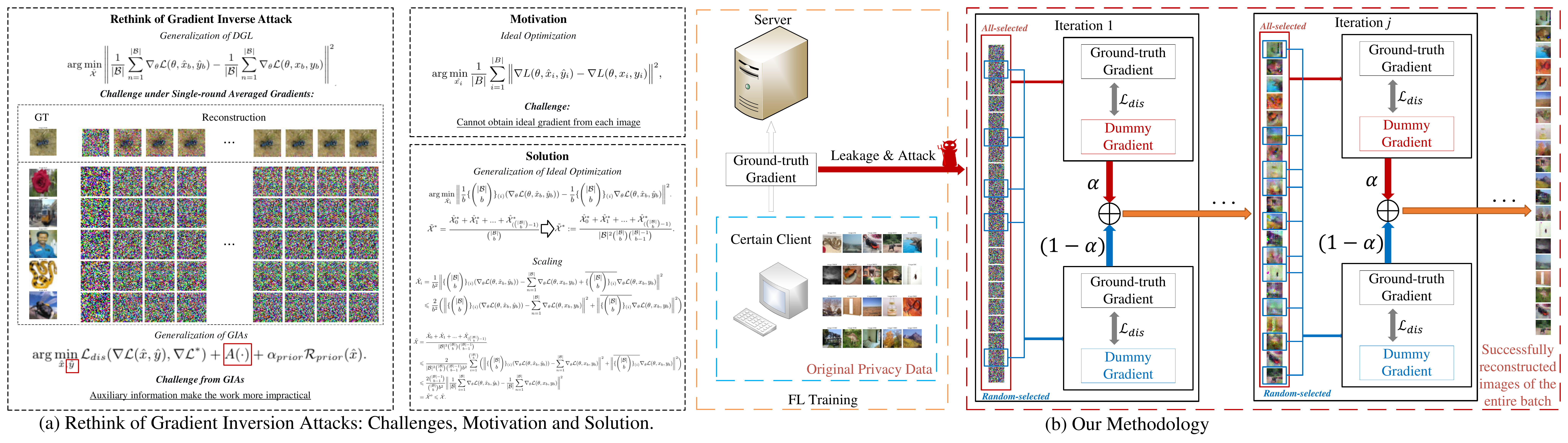}
	\caption{\textbf{MAGIA under single-round averaged gradients (SAG).} The attacker alternates all-selected and random-selected subset gradient matching to sense per-image signals from the averaged gradient and reconstruct the entire batch—without auxiliary label inference.}
	\label{fig:framework}
\end{figure*}

\textbf{Challenges and Related Works.}
The core challenge in the SAG regime is the entanglement of per-image signals within a single averaged gradient, which complicates reconstruction, as depicted in Fig.~\ref{fig:framework}(a). Pioneering work like DLG \cite{zhu2019dlg} first generalized gradient matching to batch averages but struggled with larger batches. To enhance stability, subsequent methods like IG \cite{geiping2020inverting} and GradInversion \cite{yin2021gradinversion} introduced stronger image priors and magnitude-invariant objectives. However, this progress introduced a new challenge: a dependency on auxiliary information \cite{zhao2020idlg, yu2025gi}. As illustrated in Fig.~\ref{fig:framework}(a) ('Challenge from GIAs'), many state-of-the-art attacks rely on brittle heuristics like explicit label inference ($A(\cdot)$ in the objective function), which adds assumptions and reduces practical applicability \cite{zhao2020idlg, yu2025gi}.

\textbf{Our Insight: From an Impossible Ideal to a Practical Solution.}
The limitations of prior work highlight a fundamental dilemma. The ideal scenario, as motivated in Fig.~\ref{fig:framework}(a), would be to optimize each image's gradient individually. But this is impossible, as we "cannot obtain ideal gradient from each image" from the averaged update. Our key insight is to bridge the gap between the failing whole-batch approach and the impossible single-image ideal. We propose to treat the average not as a monolith, but as a mixture that can be probed. By matching the gradients of random \emph{subsets} of the dummy variables to the global average, we can effectively "sense" the underlying per-image signals without any auxiliary information.

\textbf{The MAGIA Framework.}
Our method, \textbf{MAGIA}, operationalizes this insight. As shown in our methodology flowchart (Fig.~\ref{fig:framework}(b)), MAGIA alternates between matching the whole batch ("All-selected") and a randomly sampled subset ("Random-selected"). This mixed objective is stabilized by two key technical contributions derived from our subset-probing formulation: (1) a closed-form \textbf{combinatorial rescaling} that ensures the objective is consistent across different subset sizes, and (2) a lightweight \textbf{temporal-smoothness} term that stabilizes the stochastic search. This creates a closed-loop system where challenges identified in prior works are directly addressed by our proposed solution.

\textbf{Contributions.}
Our contributions are: (1) A novel, SAG-specific formulation that uses subset probing to extract per-image cues from an averaged gradient. (2) A compact, scale-consistent objective function that gracefully recovers standard gradient matching as a special case. (3) A practical and lightweight solver that generalizes DLG-style attacks to larger batches under SAG without label inference. (4) Systematic experiments that validate our method's effectiveness and align with calls for more realistic GIA evaluations.

\section{Methodology}\label{methodology}

\subsection{Motivation: Sensing Internal Signals in Averaged Gradients}
\label{subsec:motivation}

\textbf{From single-sample success to SAG limitations.}
State-of-the-art GIAs have shown that, given \emph{per-sample} gradients, one can reliably reconstruct inputs: DLG optimizes dummy data and labels to match shared gradients, and iDLG further extracts the true label analytically for cross-entropy with one-hot targets, stabilizing single-image recovery. However, these successes do not directly translate to the \emph{single-round averaged gradient} (SAG) regime, where the attacker only observes the batch \emph{average} and not any individual gradient. In SAG, treating the average as a monolith blurs which instances dominate the signal, making multi-sample reconstruction substantially harder.

\textbf{Avoiding fragile auxiliary heuristics.}
Recent methods often lean on auxiliary information---most prominently, explicit label inference or label-count heuristics---to boost reconstruction. While effective, this dependency makes the attack \textbf{brittle} and highly specific to model architectures (e.g., reliance on last-layer structure) or loss functions. Our goal is to eschew such fragile heuristics and instead develop a more fundamental method that extracts information \emph{directly} from the averaged gradient's internal structure.

\textbf{Core question: can we \emph{sense} per-image signals from an average?}
Revisiting DLG in the SAG setting, the fundamental gap is the lack of access to each sample's ground-truth gradient; we only have the batch mean. If we could endow the optimizer with a way to \emph{sense} or approximate the gradient contribution of a \emph{certain} sample (or a small subset) \emph{through} the average, we could recover DLG's effectiveness for multiple images under SAG \emph{without} adding external side information.

\textbf{Idealized target as a design lens.}
To formalize this, we posit an idealized oracle. Imagine that for any subset size $b\!\in\!\{1,\dots,B\}$, this oracle could reveal the optimal objective value from a perfect attack on just that subset. Averaging these theoretical optimal values over \emph{all} ${B \choose b}$ subsets would yield a superior optimization target:
\begin{equation}
\label{eqn:~X*}
\mathbb{E}(\hat{\mathcal{X}}^{\ast})
\;:=\;
\frac{
\hat{\mathcal{X}}^{\ast}_{0}+\hat{\mathcal{X}}^{\ast}_{1}+\cdots+\hat{\mathcal{X}}^{\ast}_{({B \choose b}-1)}
}{
{B \choose b}
}\,.
\end{equation}
While this oracle is purely a thought experiment, it provides a powerful \textbf{design lens}. It suggests that the key to unlocking the averaged gradient lies in its combinatorial substructure. Our method, therefore, is not to build this oracle, but to use it as a starting point to mathematically derive a \textbf{practical surrogate objective}. By re-examining the standard DLG loss through this combinatorial lens, we derive a provably tighter, scale-consistent objective that implicitly forces the optimization to respect these underlying subset statistics, thereby sensing the per-sample signals.

\subsection{Detailed Method}\label{subsec:detailed}

\textbf{Deriving the Objective via Combinatorial Analysis}
Our goal is to improve upon the standard DLG objective, which we define as $\hat{\mathcal{X}} := \|\frac{1}{|\mathcal{B}|} \sum \nabla_{\theta}\mathcal{L}_i - g^*\|^2$. We leverage the combinatorial insight from our motivation to derive a provably tighter optimization objective. To this end, we introduce $\mathcal{S}$ as our unified \emph{subset size selector}, a parameter that can be a fixed constant or vary over time during optimization.

Our derivation begins with the idealized, rescaled combinatorial target, $\tilde{\mathcal{X}}^{\ast}$, which provides a properly scaled, theoretical benchmark for the best achievable loss based on subset analysis:
\begin{equation}
\label{eq:oracle-rescaled}
\tilde{\mathcal{X}}^{\ast}
\;:=\;
\frac{\mathbb{E}_{\,\mathcal{S}\sim\mathcal{U}_\mathcal{S}}\!\big[\hat{\mathcal{X}}^{\ast}(\mathcal{S})\big]} {|\mathcal{B}|^{2}\binom{|\mathcal{B}|-1}{\mathcal{S}-1}}\! .
\end{equation}
While $\tilde{\mathcal{X}}^{\ast}$ itself is incomputable, it serves as the formal starting point for our derivation. By applying the triangle inequality to the gradient differences within each subset, we can establish a computable upper bound for this idealized target. The core steps of this derivation are as follows:
\begin{equation}
\label{eq:derivation-skeleton}
\begin{aligned}
\tilde{\mathcal{X}}^{\ast}
&\;\propto\;
\mathbb{E}_{\,\mathcal{S}} \!\left[ \min \left\| \frac{1}{\mathcal{S}}\sum_{j \in \mathcal{S}} \nabla\mathcal{L}_j - \frac{1}{\mathcal{S}}\sum_{j \in \mathcal{S}} \nabla\mathcal{L}^*_j \right\|^2 \right]
\\[0.5em]
&\;\le\;
\underbrace{ \frac{2 \binom{|\mathcal{B}|-1}{\mathcal{S}-1}}{\binom{|\mathcal{B}|}{\mathcal{S}}\, \mathcal{S}^{2}} \Bigg\| \frac{1}{|\mathcal{B}|} \sum_{i=1}^{|\mathcal{B}|} \nabla_{\theta}\mathcal{L}_i - g^* \Bigg\|^{2} }_{\text{A computable upper bound derived from combinatorial analysis}}
\end{aligned}
\end{equation}
The proof is completed through the application of the triangle inequality as Appendix \ref{appendix:a}. This process yields a new, practical objective function, which we term $\tilde{\mathcal{X}}^{o}$, that is directly derived from our theoretical ideal. This new objective is defined by the final expression in Eqn.~\eqref{eq:derivation-skeleton}.

The significance of this result is that our derived objective is provably a tighter bound than the original DLG objective:
\begin{equation}
\label{eq:tighter-lb}
\tilde{\mathcal{X}}^{o}\ \le\ \hat{\mathcal{X} = \Bigg\| \frac{1}{|\mathcal{B}|} \sum_{i=1}^{|\mathcal{B}|} \nabla_{\theta}\mathcal{L}_i - g^* \Bigg\|^{2}}.
\end{equation}
Here, $\tilde{\mathcal{X}}^{o}$ is the scaled batch-discrepancy we use in practice. Eqn.~\eqref{eq:tighter-lb} demonstrates that our objective creates a more constrained optimization landscape than standard gradient matching. Notably, it gracefully collapses to the standard DLG objective when the subset size $\mathcal{S}$ equals the full batch size $|\mathcal{B}|$, while exploiting the intra-batch structure for any $\mathcal{S}\!<\!|\mathcal{B}|$.

\textbf{Momentum-based Mixed Objective} 
The tighter objective $\tilde{\mathcal{X}}^{o}$ derived in the previous section provides a superior optimization target. However, a better static target does not guarantee a more effective dynamic search. An optimizer minimizing only the global objective might converge to a \emph{spurious solution}: a reconstruction that matches the full-batch average correctly, but whose internal structure is implausible and not robust to subset variations.

To address this, we introduce a dual-objective framework that actively probes the solution's internal structure during optimization. This framework is designed to balance two competing yet complementary goals: \textbf{(1) Global Consistency:} The primary goal is to ensure the full-batch average gradient of our dummy data matches the observed gradient $g^*$. This is driven by the standard DLG loss, $\hat{\mathcal{X}}$. \textbf{(2) Local Plausibility:} To prevent spurious solutions, we introduce a stochastic regularizer. At each iteration, we "audit" the current reconstruction by computing the gradient loss on a new random subset, $\hat{\mathcal{X}}_{i}$. This term penalizes solutions that are not structurally sound across different random samplings.
We combine these two objectives in what we term a \textbf{momentum-based} approach. The global consistency term provides a stable, "inertial" direction for the optimization, while the stochastic local plausibility term injects a fresh "impulse" at each step. The hyperparameter $\alpha$ balances the influence of the stable, historical direction against the new, stochastic information, analogous to a momentum coefficient. This leads to our final mixed objective function, which is scaled by our adaptive coefficient:
\begin{equation}
\label{eq:magia-final}
\tilde{\mathcal{X}}^{f}
\;:=\;
\underbrace{\frac{2\binom{|\mathcal{B}|-1}{\mathcal{S}-1}}{\binom{|\mathcal{B}|}{\mathcal{S}}\,\mathcal{S}^{2}}}_{\text{Adaptive Coeff.}}
\underbrace{\Big(\alpha\,\hat{\mathcal{X}} \;+\; (1{-}\alpha)\,\hat{\mathcal{X}}_{i}\Big)}_{\text{Momentum-based Mixed Discrepancy}},
\end{equation}
where $\hat{\mathcal{X}}_{i}$ is the computable local surrogate for the loss on a randomly chosen subset $i$:
\begin{equation}
\label{eq:local-surrogate}
\hat{\mathcal{X}}_{i}
\;:=\;
\left\|
\frac{1}{\mathcal{S}}\!\!\!\sum_{j\in \{\text{subset}\}_{i}}\!\!\!\nabla_{\theta}\mathcal{L}(\theta,\hat{x}_j,\hat{y}_j)
-
g^*
\right\|^{2}\!.
\end{equation}
This formulation directly operationalizes our core insight. The adaptive coefficient creates a tighter optimization landscape, and the momentum-based objective performs a robust, dynamic search within that landscape to find structurally plausible solutions.

\textbf{Image Prior Term}
In this paper, we set $\mathcal{R}_{prior}$ as total variation \cite{tv}, can be written as $\alpha_{\mathrm{tv}}\;\mathcal{R}_{\mathrm{tv}}(\hat{X}_{t})$. Total variation is a mathematical method for measuring the smoothness of an image by evaluating the sum of its gradient magnitudes and widely used in GIAs.

\section{Experiments}\label{experiments}

\begin{figure*}[t!]
	\centering
	\includegraphics[width=0.98\linewidth]{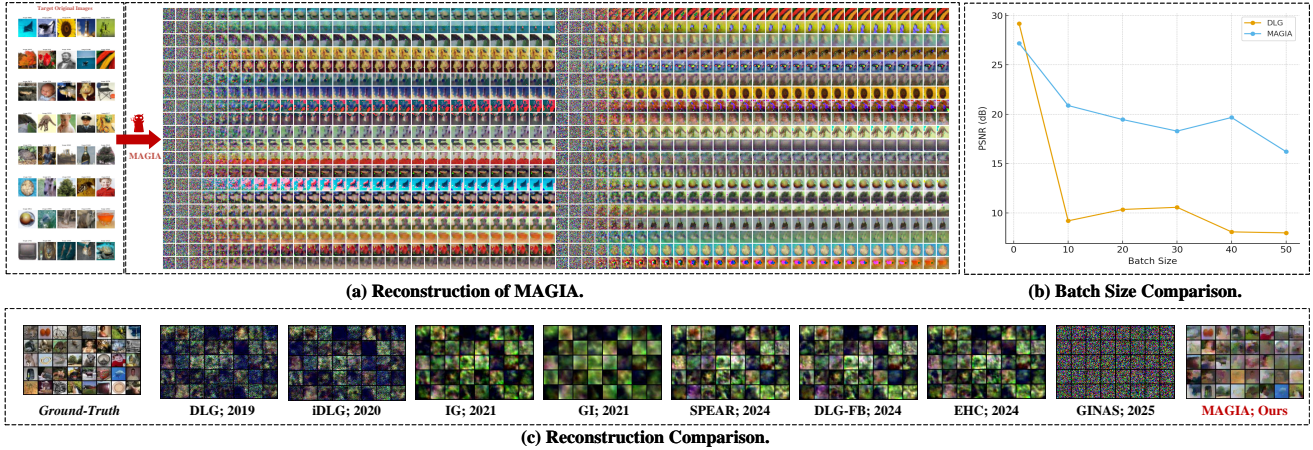}
	\caption{MAGIA reconstruction performance. (a) Qualitative results ($|\mathcal{B}|=40$). (b) PSNR comparison across batch sizes. (c) Qualitative comparison against advance baselines.}
	\label{fig:recons}
\end{figure*}

\begin{table*}[t!]
	\centering
	\setlength{\tabcolsep}{1mm}
	\caption{Comparison of Reconstruction Metrics for Different Datasets. All data is rounded to two decimal places.\label{tab:comparison}}
	\small
	\begin{tabular}{cc|cccc|cccc}
		\toprule
		\multicolumn{2}{c|}{$|\mathcal{B}|=40$} & \multicolumn{2}{c|}{FEMNIST\_byclass}  & \multicolumn{2}{c|}{EMINIST-L}  & \multicolumn{2}{c|}{SVHN} & \multicolumn{2}{c}{CIFAR-100} \\
		\cmidrule(r){3-6} \cmidrule(r){7-10}
		\multicolumn{2}{c|}{Baselines\centering} & $RMSE(\downarrow)$ & $PSNR(\uparrow)$ & $RMSE(\downarrow)$ & $PSNR(\uparrow)$& $RMSE(\downarrow)$ & $PSNR(\uparrow)$& $RMSE(\downarrow)$ & $PSNR(\uparrow)$ \\
		\midrule
		\multicolumn{2}{c|}{DLG} & 0.16 ± 0.04 & 16.18 ± 2.46 & 0.38 ± 0.04 & 8.47 ± 0.99 & 0.29 ± 0.05 & 10.96 ± 1.67 & 0.48 ± 0.03 & 6.48 ± 0.57 \\
            \multicolumn{2}{c|}{iDLG}    & 0.46 ± 0.03 & 7.14 ± 0.80   & 0.48 ± 0.04 & 6.46 ± 0.71           & 0.47 ± 0.07 & 6.71 ± 1.09   & 0.42 ± 0.13 & 8.26 ± 2.30   \\
\multicolumn{2}{c|}{IG} & 0.31 ± 0.03 & 9.64 ± 2.44 & 0.24 ± 0.05 & 12.59 ± 1.78 & 0.16 ± 0.05 & 16.25 ± 2.75 & 0.15 ± 0.06 & 16.82 ± 3.23 \\
\multicolumn{2}{c|}{GI} & 0.21 ± 0.02 & 13.61 ± 0.87 & 0.27 ± 0.03 & 11.53 ± 1.15 & 0.15 ± 0.06 & 16.84 ± 3.06 & 0.23 ± 0.07 & 13.36 ± 3.00 \\
\multicolumn{2}{c|}{SPEAR} & 0.15 ± 0.04 & 16.75 ± 2.32 & 0.28 ± 0.04 & 11.20 ± 1.11 & 0.45 ± 0.03 & 6.88 ± 0.61 & 0.20 ± 0.05 & 14.43 ± 2.24 \\
\multicolumn{2}{c|}{DLG-BF}  & 0.28 ± 0.07 & 11.32 ± 2.03 & 0.35 ± 0.04 & 9.13 ± 1.06 & 0.19 ± 0.06 & 15.17 ± 3.01 & 0.48 ± 0.03 & 6.46 ± 0.54 \\
\multicolumn{2}{c|}{EHC}  & 0.15 ± 0.02 & 18.00 ± 5.21 & 0.30 ± 0.04 & 10.65 ± 1.15 & 0.19 ± 0.07 & 15.11 ± 3.17 & 0.20 ± 0.05 & 14.43 ± 2.08 \\
\multicolumn{2}{c|}{GINAS}  & 0.20 ± 0.07 & 14.67 ± 3.24 & 0.31 ± 0.04 & 10.14 ± 1.10 & 0.18 ± 0.06 & 15.34 ± 2.74 & 0.23 ± 0.05 & 12.85 ± 1.85 \\
		\multicolumn{2}{c|}{\textbf{MAGIA}}  & $\mathbf{0.13\pm0.06}$ & $\mathbf{19.17\pm4.76}$ & $\mathbf{0.23\pm0.20}$ & $\mathbf{18.22\pm3.12}$  & $\mathbf{0.14\pm0.04}$ & $\mathbf{19.59\pm2.95}$ & $\mathbf{0.11\pm0.04}$ & $\mathbf{19.67\pm2.95}$ \\
		\bottomrule
	\end{tabular}
\end{table*}

\begin{figure}[t!]
	\centering
	\includegraphics[width=0.98\linewidth]{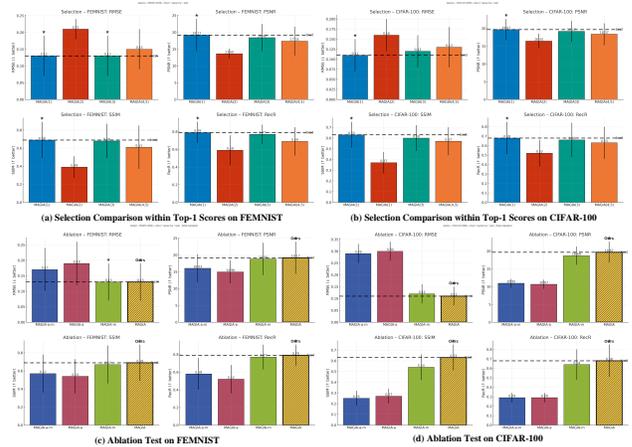}
	\caption{Analysis of MAGIA's key components and hyperparameters. (Top Row) Performance comparison of different selection strategies for the subset size $\mathcal{S}$. (Bottom Row) Ablation study results on FEMNIST and CIFAR-100, showing the impact of the adaptive coefficient (-m) and the momentum term (-a).}
	\label{fig:selaabla}
\end{figure}

\textbf{Setup.} We conducted experiments with DLG \cite{zhu2019dlg}, iDLG \cite{zhao2020idlg}, IG \cite{geiping2020inverting}, GI \cite{yin2021gradinversion}, SPEAR \cite{dimitrov2024spear}, DLG-BF \cite{leite2024federated}, GINAS \cite{yu2025gi} and EHC \cite{wang2024towards} on FEMNIST-class (62-class) \cite{caldas2018leaf}, EMNIST-L (26-class) \cite{cohen2017emnist}, SVHN (10-class) \cite{netzer2011reading}, CIFAR-100 (100-class) \cite{krizhevsky2009learning}. Details shown in Appendix \ref{appendix:b}.

\textbf{Inversion \& Reconstruction.} We evaluate MAGIA against eight baseline methods on four standard datasets, with results summarized in Table~\ref{tab:comparison} and Fig.~\ref{fig:recons}. The quantitative results show that MAGIA substantially outperforms all baselines across all tested datasets. For instance, on FEMNIST and CIFAR-100 with a large batch size of $|\mathcal{B}|=40$, MAGIA achieves high-fidelity reconstructions with PSNR scores of $19.17$ and $19.67$, respectively, while competing methods largely fail to produce effective results. This quantitative superiority is visually confirmed in Fig.~\ref{fig:recons}(c), where MAGIA's reconstructions are significantly more recognizable than those from other state-of-the-art methods. Furthermore, Fig.~\ref{fig:recons}(b) highlights MAGIA's robustness to larger batch sizes; while performance degrades for all methods as batch size increases, MAGIA maintains a consistently and significantly higher PSNR than the baseline DLG, demonstrating the effectiveness of our subset-probing approach.

\textbf{Selection \& Ablation.} The analysis of subset size $\mathcal{S}$ (Fig.~\ref{fig:selaabla}, Top Row) demonstrates the framework's robustness, as strong results are maintained across various strategies, including schedules that are time-varying with the optimization iteration. The ablation study (Bottom Row) confirms the full MAGIA model achieves the best performance, validating the necessity of both the adaptive coefficient and the momentum-based objective.

\section{Conclusion}
We introduced MAGIA, a SAG-specific framework that improves reconstruction quality without label inference using a tighter, rescaled objective and a momentum-mixed loss. Our formulation provides a robust generalization of standard methods, demonstrating strong performance across diverse datasets and batch sizes. Deatils shown in Appendix \ref{appendix:c}.

\clearpage
\bibliographystyle{IEEEtran}
\bibliography{refs}

\clearpage
\appendix
\input{appendixA}
\clearpage
\input{appendixB}
\clearpage
\input{appendixC}

\end{document}

%% file: appendixA.tex
\section{Detailed Method with Deduction}\label{appendix:a}

\subsection{Rethink: from DLG to GIA}

The concept of deep leakage from gradients was first introduced by \cite{zhu2019dlg}, which laid the groundwork for understanding privacy vulnerabilities in gradient-based methods. By extending their application scenario from single samples to batch training, the optimization objective for deep leakage can be generalized as follows:

\begin{equation}
	\label{eqn:X*}
	\begin{aligned}
		&\hat{\mathcal{X}}^*=	\\
		&\arg \min_{\hat{\mathcal{X}}}\left\| \frac{1}{|\mathcal{B}|} \sum_{n=1}^{|\mathcal{B}|} \nabla_{\theta} \mathcal{L}(\theta, \hat{x}_b, \hat{y}_b) - \frac{1}{|\mathcal{B}|} \sum_{n=1}^{|\mathcal{B}|} \nabla_{\theta} \mathcal{L}(\theta, x_b, y_b) \right\|^2.
	\end{aligned}
\end{equation}
where $\mathcal{B}$ is the batch size; $\theta$ is the net weights; $\hat{x}$ and $\hat{y}$ represent dummy data and dummy labels, respectively. Adversaries generate dummy gradients using these variables, iteratively updating them to minimize the difference between the dummy and target gradients. As optimization progresses, the dummy gradient increasingly approximates the target gradient, ultimately exposing the ground-truth data.

Building on this foundation, the general optimization objective for GIAs can be expressed as:

\begin{equation}
	\label{eqn:gia}
	\arg \min_{\hat{x},\hat{y}}\mathcal{L}_{dis}(\nabla\mathcal{L}(\hat{x},\hat{y}),\nabla\mathcal{L}^*) + A(\cdot) + \alpha_{prior}\mathcal{R}_{prior}(\hat{x}).
\end{equation}
where $\mathcal{L}_{dis}$ is the loss to regularize dummy gradient and target, such as $L2-$norm and cosine similarity; $A(\cdot)$ is the certain penalization from auxiliary information; $\mathcal{R}_{prior}$ is image prior regularization. Once the optimization is converge, the dummy data generated by adversaries reveals the hidden information in the ground-truth gradient.

\subsection{Motivation: Exploiting Intra-Batch Gradient Dynamics}

Existing Gradient Inversion Attacks (GIAs) have demonstrated the potential to reconstruct private data from shared gradients in Federated Learning (FL) \cite{zhu2019dlg, geiping2020inverting, zhao2020idlg}. However, many contemporary approaches exhibit significant limitations, hindering their effectiveness in practical scenarios. Firstly, reliance on auxiliary information, such as label counts or specific network architectures, introduces computational overhead and restricts applicability \cite{dimitrov2022dlf, lre1}. Secondly, and more critically, most existing GIAs implicitly treat the aggregated gradient of an entire batch as a monolithic entity \cite{zhu2019dlg, lti}. They typically optimize a dummy input by minimizing the distance to the target batch gradient, effectively overlooking the rich structural information embedded within the batch gradient itself.

Our core insight was to recognize that while direct single-image gradient recovery is impossible under whole-batch uploads, partial-batch gradients—computed over random subsets of the batch—serve as intermediate approximations that can be sampled in practice. By invoking the triangle inequality and summing over all combinatorial subsets $\mathcal{S}_i$ of size $b$, we derive a surrogate loss
\begin{equation}
	\arg \min_{\hat{x_i}} \frac{1}{ {\binom{|B|}{b}}} \sum_{i=1}^{\binom{|B|}{b}}
	\Big\|   \nabla L(\theta, \hat{x}_i, \hat{y}_i)
	-  \nabla L(\theta, x_i, y_i) \Big\|^2,
\end{equation}
whose lower bound exceeds the ideal single-image objective but remains below the naive batch-level objective. This formulation explicitly encodes the relationship between partial-batch and whole-batch gradients, enabling us to exploit intra-batch dynamics to extend gradient inversion attacks to larger batch sizes without additional label information.

To investigate this hypothesis in a controlled setting, we focus on the SAG framework which introduced in main text. Understanding leakage in this fundamental setting is crucial for developing insights applicable to more complex FL scenarios.

Therefore, our motivation is to develop a GIA framework, MAGIA, that directly addresses the limitations of prior work by:
\begin{itemize}
	\item Eliminating the need for label reconstruction.
	\item Explicitly bridging the gap between local (partial-batch) and global (whole-batch) gradient information.
	\item Exploiting estimated local optima to refine the global reconstruction process.
\end{itemize}
By tackling the overlooked dimension of intra-batch dynamics, MAGIA aims to provide a more powerful tool for understanding gradient leakage and ultimately informing the development of stronger privacy defenses in FL.

\begin{algorithm}[t!]  
	\caption{Proposed MAGIA}  
	\label{alg:proposed}  
	\textbf{Input}: $\nabla\mathcal{L}^*$\\
	\textbf{Parameters}: $\hat{x}$, $\hat{y}$, $N$, $\mathcal{S}$, $\alpha$, $\beta$, $I$, $\alpha_{prior}$\\
	\textbf{Output}: $\hat{x}$  
	\begin{algorithmic}  
		\FOR{$i\leqslant I$}  
		\STATE Update $\mathcal{S}$ as experiment of effect of $\mathcal{S}$.
		\STATE Randomly sampling as $\{\binom{|\mathcal{B}|}{\mathcal{S}}\}_{(i)}$. 
		\STATE Calculate $\mathcal{L}_o=$$\left\| \frac{1}{|\mathcal{B}|}\nabla_{\theta}\mathcal{L}(\theta,\hat{x}_b,\hat{y}_b)-\frac{1}{|\mathcal{B}|}\nabla_{\theta}\mathcal{L}(\theta,{x}_b,{y}_b)\right\|^2$.
		\STATE Calculate $\mathcal{L}_r=$\\
		\ \ \ \ \ \ \ $\left\|  \frac{1}{\mathcal{S}}\{\binom{|\mathcal{B}|}{\mathcal{S}}\}_{(i)}(\nabla_{\theta} \mathcal{L}(\theta, \hat{x}_n, \hat{y}_n)) - \frac{1}{|\mathcal{B}|}\nabla_{\theta}\mathcal{L}(\theta, {x}_n, {y}_n) \right\|^2$.
		\STATE Update and backward \\ $\mathcal{L}_{total}=\frac{2\binom{|\mathcal{B}|-1}{\mathcal{S}-1}}{\binom{|\mathcal{B}|}{\mathcal{S}}\mathcal{S}^2}(\alpha\mathcal{L}_o+(1-\alpha)\mathcal{L}_r)+\alpha_{prior}\mathcal{R}_{prior}(\hat{x})$
		\ENDFOR 
	\end{algorithmic}  
\end{algorithm} 



%
\subsection{Adaptation Coefficient} 

Within the GIA framework, we propose MAGIA, a novel momentum-based adaptation correction method to enhance gradient inversion in SAG. A key feature of our method is the introduction of a variable $\mathcal{S}$, which serves as a bridge between local and global optimal solutions, enabling more precise constraints. In this paper, we set $\mathcal{R}_{prior}$ as total variation. 

Eqn.\ref{eqn:X*} expresses the global optimal solution of the entire dummy batch to ground-truth gradient in SAG using L2 norm. Considering the local optimal solution $\hat{\mathcal{X}}^*_i$ of specific image combination in batch, we can express it as follows:


\begin{equation}  
	\label{eqn:x_i^x}  
	\begin{split}  
		&\arg\min_{\hat{\mathcal{X}}_i} \bigg\| \frac{1}{b} \{\binom{|\mathcal{B}|}{b}\}_{(i)} (\nabla_{\theta} \mathcal{L}(\theta, \hat{x}_b, \hat{y}_b)) - \\
		&\quad\quad\quad\quad\quad\quad\quad\quad\quad\frac{1}{b} \{\binom{|\mathcal{B}|}{b}\}_{(i)} \nabla_{\theta} \mathcal{L}(\theta, \hat{x}_b, \hat{y}_b) \bigg\|^2.  
	\end{split}  
\end{equation}

where $b\in\{1,2,...,|\mathcal{B}|\}$ is the selected number of combination, means partial batch size;  $\binom{|\mathcal{B}|}{b}$ is the number of combinations; $\{\binom{|\mathcal{B}|}{b}\}_{(i)}$ shows the $i$-th combination; $\{\binom{|\mathcal{B}|}{b}\}_{(i)} \nabla_{\theta} \mathcal{L}(\theta, \hat{x}_b, \hat{y}_b)$ shows features and labels from partial batch that $(\cdot)_b\in\{\binom{|\mathcal{B}|}{b}\}_{(i)}$. However, we cannot obtain the optimal ground-truth gradient directly from client's uploading gradient. To solve this dilemma and utilize local optimal solution to regularize GIAs' performance, we derive an optimization objective that is tighter than Eqn.\ref{eqn:X*} from the full set of local optimal solutions. Specifically, we assume that $\mathcal{L}$ is non-negative convex and introduce an optimization objective from Eqn.\ref{eqn:~X*}:







\begin{equation}
	\label{eqn:~X*}
	\tilde{\mathcal{X}}^*:=\frac{\hat{\mathcal{X}}^*_0+\hat{\mathcal{X}}^*_1+...+\hat{\mathcal{X}}^*_{(\binom{|\mathcal{B}|}{b}-1)}}{|\mathcal{B}|^2\binom{|\mathcal{B}|}{b}\binom{|\mathcal{B}|-1}{b-1} }.
\end{equation}
The fundamental idea behind Eqn.\ref{eqn:~X*} is to derive a tighter optimization objective from $MEAN(\hat{\mathcal{X}}^*_i)$. Specifically, by considering the optimization of global combinations and the local optimization achieved by reselecting a sample and recomputing the combination, we have designed Eqn.\ref{eqn:~X*}. This approach reduces the gap between the global optimal solution and the local one, thereby further enhancing the performance of GIAs. In order to align the optimization goals of a partial batch with the optimization goals of the whole batch, we need to scale Eqn.\ref{eqn:x_i^x} for each local optimal objective.

\begin{equation}  
	\label{eqn:x_i}  
	\begin{split}  
		& \hat{\mathcal{X}}_i = \frac{1}{b^2} \bigg\| \{\binom{|\mathcal{B}|}{b}\}_{(i)} (\nabla_{\theta} \mathcal{L}(\theta, \hat{x}_b, \hat{y}_b)) - \sum_{n=1}^{|\mathcal{B}|} \nabla_{\theta} \mathcal{L}(\theta, x_b, y_b) + \\
		&\ \ \ \ \ \ \ \ \ \ \ \ \ \ \ \ \ \ \ \ \ \ \ \ \ \ \ \ \ \ \ \ \ \ \ \ \ \ \  \quad \overline{\{\binom{|\mathcal{B}|}{b}\}_{(i)}} \nabla_{\theta} \mathcal{L}(\theta, x_b, y_b) \bigg\|^2 \\
		& \leqslant \frac{2}{b^2} \bigg( \bigg\| \{\binom{|\mathcal{B}|}{b}\}_{(i)} (\nabla_{\theta} \mathcal{L}(\theta, \hat{x}_b, \hat{y}_b)) - \sum_{n=1}^{|\mathcal{B}|} \nabla_{\theta} \mathcal{L}(\theta, x_b, y_b) \bigg\|^2 + \\
		&\ \ \ \ \ \ \ \ \ \ \ \ \ \ \ \ \ \ \ \ \ \ \ \ \ \ \ \ \ \ \ \ \ \  \quad \bigg\| \overline{\{\binom{|\mathcal{B}|}{b}\}_{(i)}} \nabla_{\theta} \mathcal{L}(\theta, x_b, y_b) \bigg\|^2 \bigg) \\
	\end{split}   
\end{equation}  
where $\overline{\{\binom{|\mathcal{B}|}{b}\}_{(i)}}$ is complementary set of $\{\binom{|\mathcal{B}|}{b}\}_{(i)}$.
In relation to Eqn.\ref{eqn:~X*}, we can get the following objectives:

\begin{equation}  
	\label{eqn:io}  
	\begin{split}  
		& \tilde{\mathcal{X}} = \frac{\hat{\mathcal{X}}_0 + \hat{\mathcal{X}}_1 + ... + \hat{\mathcal{X}}_{(\binom{|\mathcal{B}|}{b}-1)}}{|\mathcal{B}|^2 \binom{|\mathcal{B}|}{b} \binom{|\mathcal{B}|-1}{b-1}} \\
		& \leqslant \frac{2}{|\mathcal{B}|^2 \binom{|\mathcal{B}|}{b} \binom{|\mathcal{B}|-1}{b-1} b^2} \sum_{i=1}^{\binom{|\mathcal{B}|}{b}} \bigg( \bigg\| \{\binom{|\mathcal{B}|}{b}\}_{(i)} (\nabla_{\theta} \mathcal{L}(\theta, \hat{x}_b, \hat{y}_b)) - \\
		&\  \quad \quad\quad \sum_{n=1}^{|\mathcal{B}|} \nabla_{\theta} \mathcal{L}(\theta, x_b, y_b) \bigg\|^2 + \bigg\| \overline{\{\binom{|\mathcal{B}|}{b}\}_{(i)}} \nabla_{\theta} \mathcal{L}(\theta, x_b, y_b) \bigg\|^2 \bigg) \\
		& \leqslant \frac{2 \binom{|\mathcal{B}|-1}{b-1}}{\binom{|\mathcal{B}|}{b} b^2} \bigg\| \frac{1}{|\mathcal{B}|} \sum_{i=1}^{|\mathcal{B}|} \nabla_{\theta} \mathcal{L}(\theta, \hat{x}_b, \hat{y}_b) - \frac{1}{|\mathcal{B}|} \sum_{n=1}^{|\mathcal{B}|} \nabla_{\theta} \mathcal{L}(\theta, x_b, y_b) \bigg\|^2 \\
		& = \tilde{\mathcal{X}}^o \leqslant \hat{\mathcal{X}}.  
	\end{split}  
\end{equation}  
Eqn.\ref{eqn:io} shows that $\tilde{\mathcal{X}^o}$ is a tighter regularization compared with DLG's optimization objective. Hence, we optimize $\tilde{\mathcal{X}^o}$ as our initial objective $\mathcal{L}_{dis}$. Further, we set $b$ as a variable, represented by $\mathcal{S}$. Its update rules are based on local iterations, as described in experiment of effect of $\mathcal{S}$. So far, we have designed an adaptive coefficient $\frac{2\binom{|\mathcal{B}|-1}{\mathcal{S}-1}}{\binom{|\mathcal{B}|}{\mathcal{S}}\mathcal{S}^2}$ of the local optimal solution obtained in a partial batch.

\subsection{Momentum-based Penalized Term}

Although we achieved a tighter optimization objective and an adaptive parameter by introducing a local optimal solution, we did not get a suitable improvement for $\mathcal{L}_{dis}$ in Eqn.\ref{eqn:gia}.
To improve the performance of our proposed method, we introduce momentum-based penalized term to regularize our $\mathcal{L}_{dis}$.
Specifically, for every iteration $i$, we employed a randomly chosen subset $\{\binom{|\mathcal{B}|}{\mathcal{S}}\}_{(i)}$ to represent a local optimal solution. Furthermore, we split $\tilde{\mathcal{X}}^o$ into momentum-based objective from Eqn.\ref{eqn:X*} and Eqn.\ref{eqn:x_i^x} with adaptive coefficient:

\begin{equation}
	\label{eqn:momentum}
	\begin{split}
		\tilde{\mathcal{X}}^f &:= \frac{2\binom{|\mathcal{B}|-1}{\mathcal{S}-1}}{\binom{|\mathcal{B}|}{\mathcal{S}}\mathcal{S}^2} (\alpha\hat{\mathcal{X}} + (1-\alpha)\hat{\mathcal{X}}_i).\\
	\end{split}
\end{equation}

However, as we have analyzed above, we cannot obtain $\hat{\mathcal{X}}_i$ directly. Therefore, we used an estimator to approximate this penalized term:

\begin{equation}
	\label{eqn:penalized}
	\begin{split}
		&\hat{\mathcal{X}}_i\approx\tilde{\mathcal{X}}^o_i\\
		&=\left\| \frac{1}{\mathcal{S}} \{\binom{|\mathcal{B}|}{\mathcal{S}}\}_{(i)} (\nabla_{\theta} \mathcal{L}(\theta, \hat{x}_b, \hat{y}_b)) - \frac{1}{|\mathcal{B}|} \sum_{n=1}^{|\mathcal{B}|} \nabla_{\theta} \mathcal{L}(\theta, \hat{x}_b, \hat{y}_b) \right\|^2.
	\end{split}
\end{equation}
So far, we have realized all the improvement of GIA performance by the local optimal solution obtained in partial batch. Our algorithm summarized in Alg.\ref{alg:proposed}.

Furthermore, taking into account Eqn.\ref{eqn:momentum}, we express the computational complexity of DLG as $\mathcal{O}(\hat{\mathcal{X}})$. The computational complexity of MAGIA can be derived as $\mathcal{O}(\tilde{\mathcal{X}}^f)=2\mathcal{O}(\hat{\mathcal{X}})\cdot\mathcal{O}(\mathcal{C})=\mathcal{O}(\hat{\mathcal{X}})$, where $\mathcal{C}$ denotes the constant-level computational overhead. It is evident that MAGIA does not elevate the computational complexity of the algorithm.

\subsection{Selection of $\mathcal{S}$}\label{sec:4.3}

For our method, the implementation of $\mathcal{S}$ realizes the complementarity between batch images and certain images reconstructed by MAGIA. Therefore, the selection of $\mathcal{S}$ is crucial for the effectiveness of the method. We primarily considered the following options:
\begin{enumerate}
\item Set $\mathcal{S}$ as a constant, where $0<\mathcal{S}\leqslant \mathcal{B}$. This is a fundamental form that ensures sufficient reconstruction of each image in the batch, regardless of how $\mathcal{S}$ is set, as long as the expectations are met.
\item Set $\mathcal{S}=\lceil{\mathcal{B}\times\frac{e+1}{E}}\rceil$, which will make $\tilde{\mathcal{X}}^f=\tilde{\mathcal{X}}^o$. The selection of $\mathcal{S}$ is related to the local iteration count, which can enhance smoothness in certain scenarios.
\item Set $\mathcal{S}=\lceil{\mathcal{B}\times\frac{e+1}{\mathcal{E}}}\rceil$, where $\mathcal{E}$ is a settable constant and $0<\mathcal{E}$. The promotion of Selection 2.
\item Set $\mathcal{S}=\lceil{\mathcal{B}\times(1-\frac{e}{E})}\rceil$. Reverse implement the coupling in Selection 2. 
\item Set $\mathcal{S}=\lceil{\mathcal{B}\times(1-\frac{e}{\mathcal{E}})}\rceil$, where $\mathcal{E}$ is a settable constant and $0<\mathcal{E}$. The promotion of Selection 4.
\end{enumerate}
In our experimental chapter, we compare the performance differences among the various schemes.

\subsection{Image Prior Term}

In this paper, we set $\mathcal{R}_{prior}$ as total variation \cite{tv}. Total variation is a mathematical method for measuring the smoothness of an image by evaluating the sum of its gradient magnitudes and widely used in GIAs.

%% file: appendixB.tex
\section{Experiments}\label{appendix:b}

In this section, we primarily discuss the content related to the experiments conducted. A deeper discussion on GIA and gradient leakage is placed in the conclusion.

\subsection{Setup}

We conducted experiments using NVIDIA GeForce RTX 4080 hardware, CUDA Driver 12.6, Driver Version 561.17, and Python 3.11 as our experimental environment. Following is more detailed setup.

\subsubsection{Datasets}

We use FEMNIST and CIFAR-100 as our comparison experiments' datasets, which are widely used in FL research and realize a variety of classification tasks. FEMNIST\_byclass is consists of 62 classes with one color channel, and CIFAR-100 is the 100-classes-dataset with three color channels. We choose LeNet as our model architecture, setting as original papers and ours as \cite{dlg}. We set batch size as 40 images and randomly sampling the batch.

We introduced the key code for the baselines within the DLG code framework. We also replicated the performance of the baselines. Notably, the original DLG paper aims to reconstruct a single image. The IG paper primarily discusses a batch size in the order of single digits. In DLF, although random batch sampling in FedAvg is considered, the maximum batch size only reaches 10. This approach does not fully align with ours. We significantly increased the upper limit of the local batch size.

\subsubsection{Metrics}

We adopt $MSE$ (Mean Squared Error), $RSME$ (Root Mean Squared Error), $PSNR$ (Peak Signal-to-Noise Ratio), $SSIM$ (Structural Similarity Index Measure) \cite{psnr} and $RecR$ (Reconstruction Rate) as our evaluation indicators. We use $Mean\pm{Standard}$ form to score the accuracy of whole batch. \cite{dlf} deems that when $PSNR>19$, means that the  reconstruction is effective.

\subsubsection{Baselines and Hyper-parameters}

We conducted comparison with DLG \cite{dlg}, IG\cite{ig}, DLFA\cite{dlf}. For IG, we set $tv=1e-5$ for FEMNIST\_byclass and $tv=1e-4$ for CIFAR-100. For DLFA, we set $\lambda_{clip}=0.2$, $\lambda_{inv}=10$, $\eta_{rec}=0.4$ on FEMNIST\_byclass and $\lambda_{clip}=10$, $\lambda_{inv}=6.075$, $\eta_{rec}=0.1$ on CIFAR-100. For MAGIA, We set momentum parameter $\alpha$ as $0.999$ and set $tv$ as 0.005. We set iteration as 300. We set L-BFGS as GIAs' optimizer. We have done batch processing of DLG which can only handle a single image.

\subsection{Detailed Experiments}

\subsubsection{Comparison Experiments}

As shown in Tab.\ref{tab:comparison}, we conducted comparative experiments with various baseline methods, demonstrating the strong performance of our proposed MAGIA. Through these comparisons, we observed the following key features of MAGIA:

\textbf{Effectiveness of the Optimization Objective.} Unlike IG, which uses cosine similarity as its optimization objective, MAGIA employs the L2 norm. Our experiments indicate that while the L-BFGS optimizer is effective for reversing gradients, combining it with cosine similarity can cause gradient explosion, negatively impacting performance. Additionally, the use of certain network configurations, such as Sigmoid activation functions instead of ReLU, can exacerbate vanishing gradient issues, further degrading accuracy. By continuing to use the L2 norm, MAGIA avoids these pitfalls and maintains a stable optimization process. We further validated this through ablation studies on the various components of our method, confirming the reliability and effectiveness of our optimization objective.

\textbf{Robustness Across Diverse Targets.} In practical applications, user data often varies greatly in complexity and diversity. To evaluate MAGIA's performance across different types of data, we conducted experiments using the single-channel FEMNIST dataset and the three-channel CIFAR-100 dataset. A common challenge for GIA methods is reconstructing images with solid-colored backgrounds, as observed in DLG, which often results in mismatched color patches. Similarly, MAGIA encounters difficulties in accurately recovering such backgrounds for both FEMNIST and CIFAR-100 datasets, as shown in the text. However, even in scenarios involving larger batch sizes, MAGIA is able to recover the content of ground-truth images with sufficient clarity for visual interpretation. We hypothesize that this limitation stems from the use of total variation as the image prior term (\(\mathcal{R}_{prior}\)) in MAGIA. Incorporating a more robust image prior term could further enhance MAGIA's ability to handle solid-colored backgrounds and improve overall performance.


\subsubsection{Effect of $\mathcal{S}$}

The biggest innovation in this paper is to reduce the gap between local and global optimal solutions by using $\mathcal{S}$. Therefore, how to choose the appropriate $\mathcal{S}$ in the iterative process becomes one of the factors that affect the accuracy of the model. Here we consider the following selection methods as Sec.\ref{sec:4.3}:
\begin{enumerate}
\item Set $\mathcal{S}$ as a constant, where $0<\mathcal{S}\leqslant \mathcal{B}$. 
\item Set $\mathcal{S}=\lceil{\mathcal{B}\times\frac{e+1}{\mathcal{E}}}\rceil$, which will make $\tilde{\mathcal{X}}^f=\tilde{\mathcal{X}}^o$. 
\item Set $\mathcal{S}=\lceil{\mathcal{B}\times\frac{e+1}{\mathcal{E}}}\rceil$, where $\mathcal{E}$ is a settable constant and $0<\mathcal{E}$. After $1\leqslant\mathcal{S}$, we set $\mathcal{S}=1$. When $\mathcal{E}=E$, effect same as (2). When $\mathcal{E}=\mathcal{B}\times E$, effect same as (1) and set $\mathcal{S}=1$. In this comparison, we set $\mathcal{E}_{(2)}=\{100,150,200,250,300,500,1000,2000,\\3000,5000,10000\}$.
\item Set $\mathcal{S}=\lceil{\mathcal{B}\times(1-\frac{e}{E})}\rceil$. 
\item Set $\mathcal{S}=\lceil{\mathcal{B}\times(1-\frac{e}{\mathcal{E}})}\rceil$, where $\mathcal{E}$ is a settable constant and $0<\mathcal{E}$. When $\mathcal{E}=E$, effect same as (4). In this comparison, we set $\mathcal{E}_{(5)}=\{300,500,1000,1500,2000,\\3000,5000,10000\}$.
\end{enumerate} 


\begin{figure}[!t]
	\centering
	\subfloat[]{\includegraphics[width=0.13\textwidth]{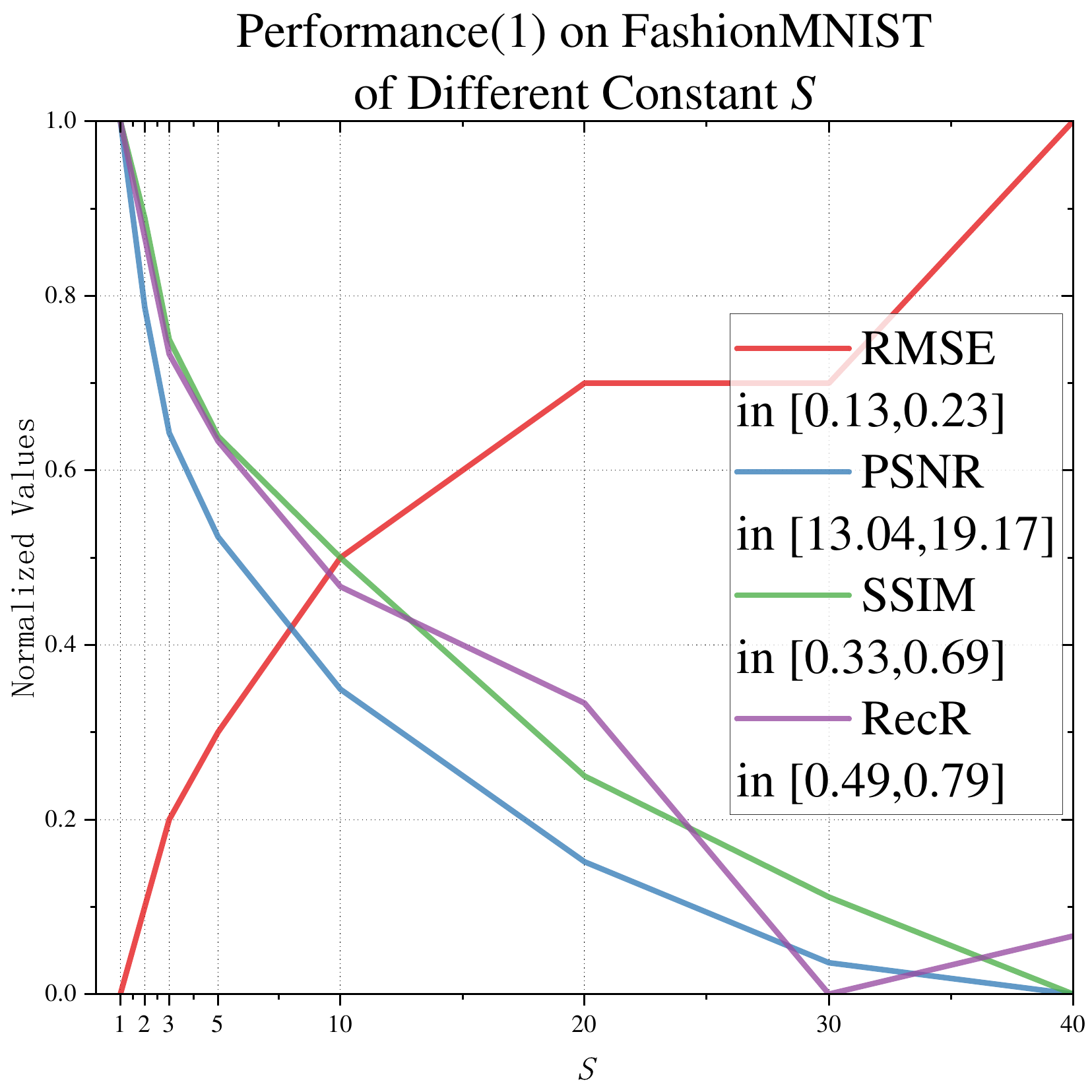}%
		\label{fig:constant_fm}}
	\hfil
	\subfloat[]{\includegraphics[width=0.13\textwidth]{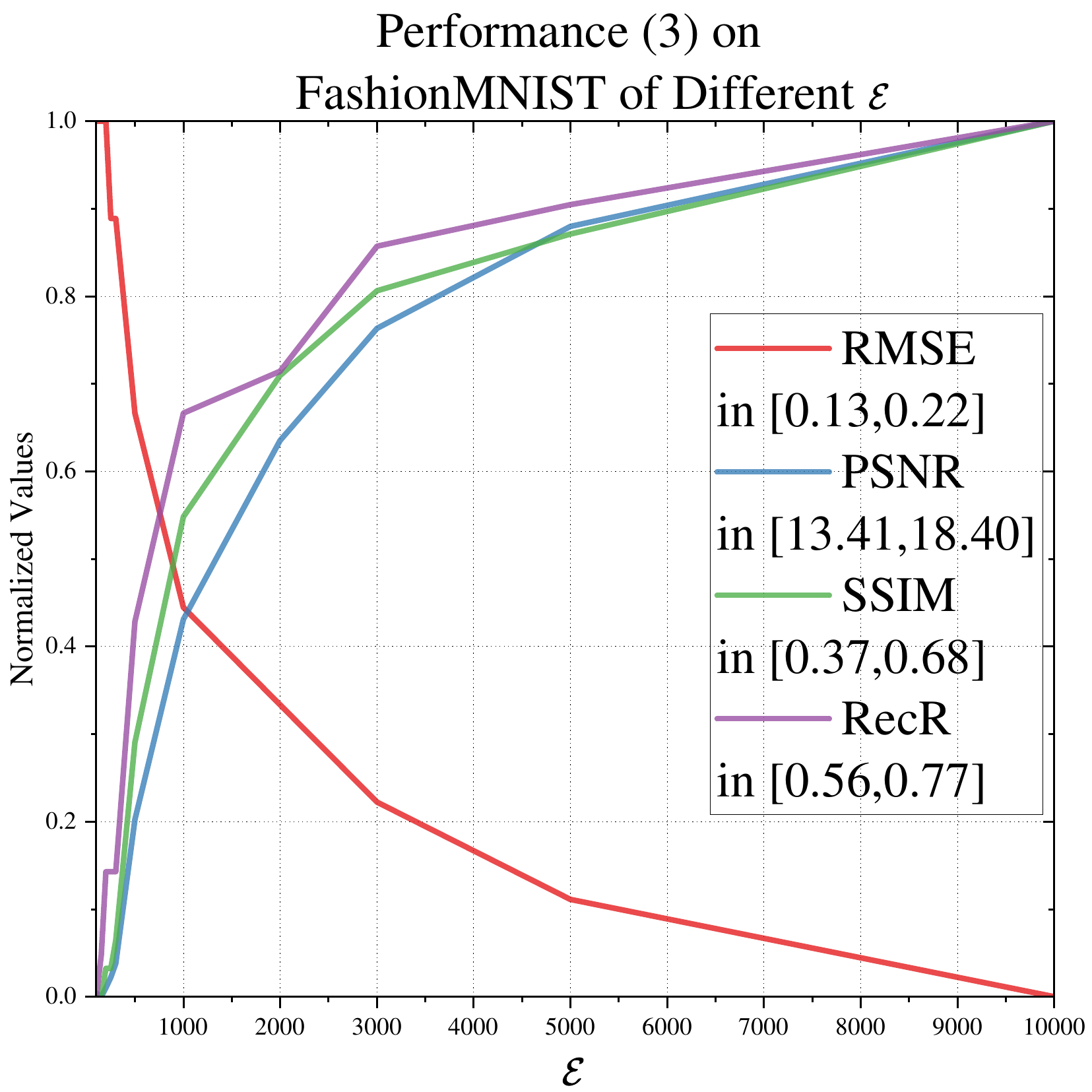}%
		\label{fig:E+_fm}}
	\hfil
	\subfloat[]{\includegraphics[width=0.13\textwidth]{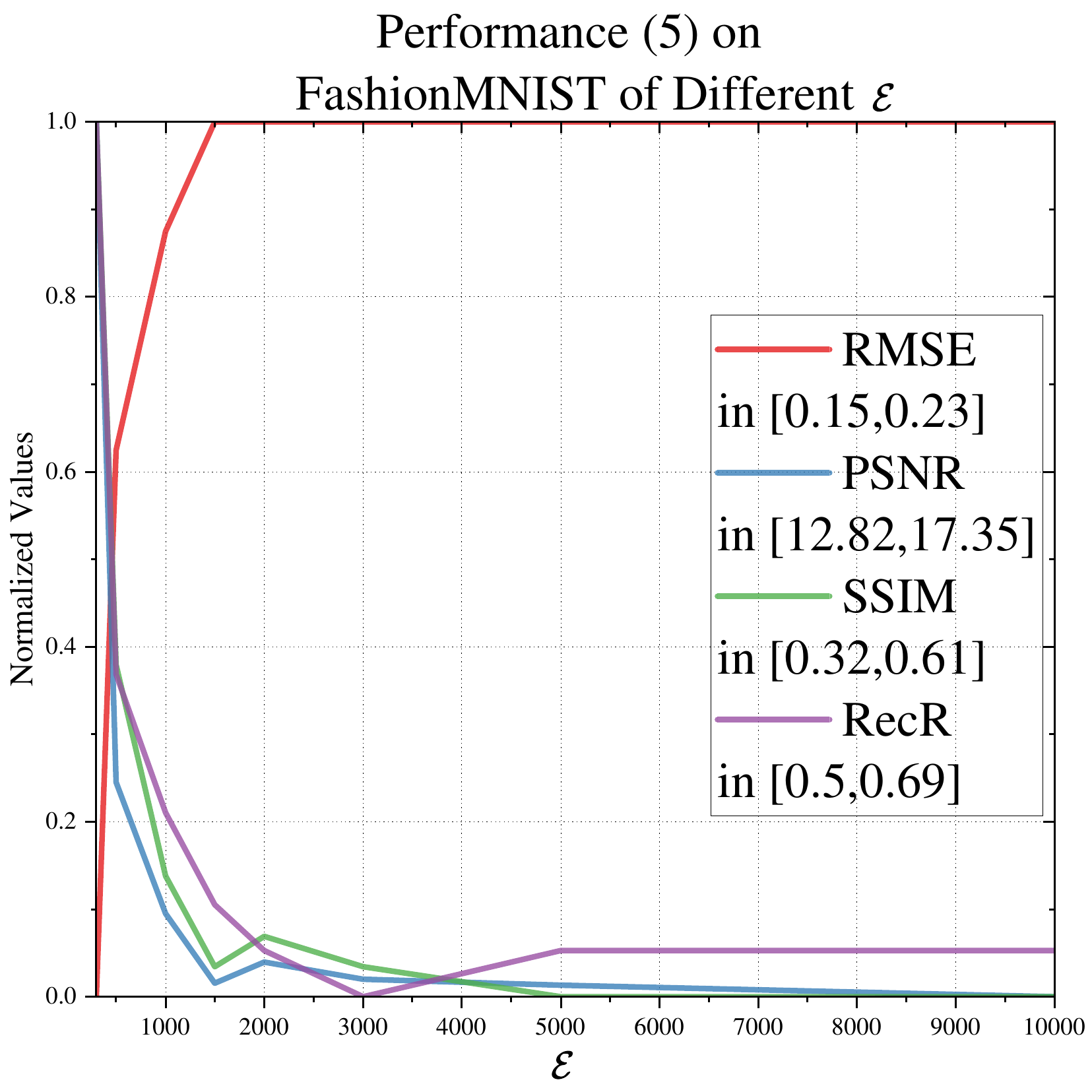}%
		\label{fig:E-_fm}}
	\hfil\\
	\subfloat[]{\includegraphics[width=0.13\textwidth]{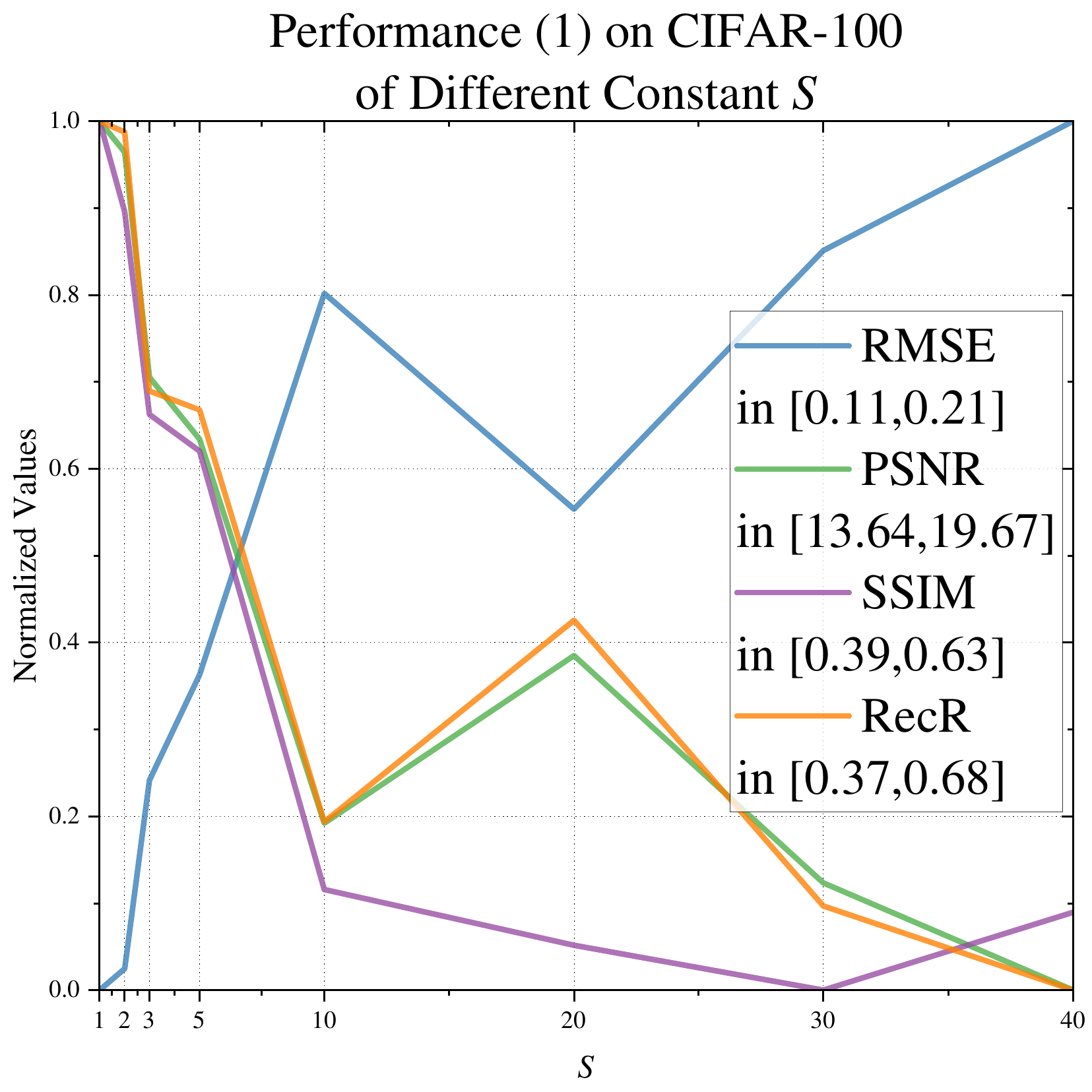}%
		\label{fig:constant_c100}}
	\hfil
	\subfloat[]{\includegraphics[width=0.13\textwidth]{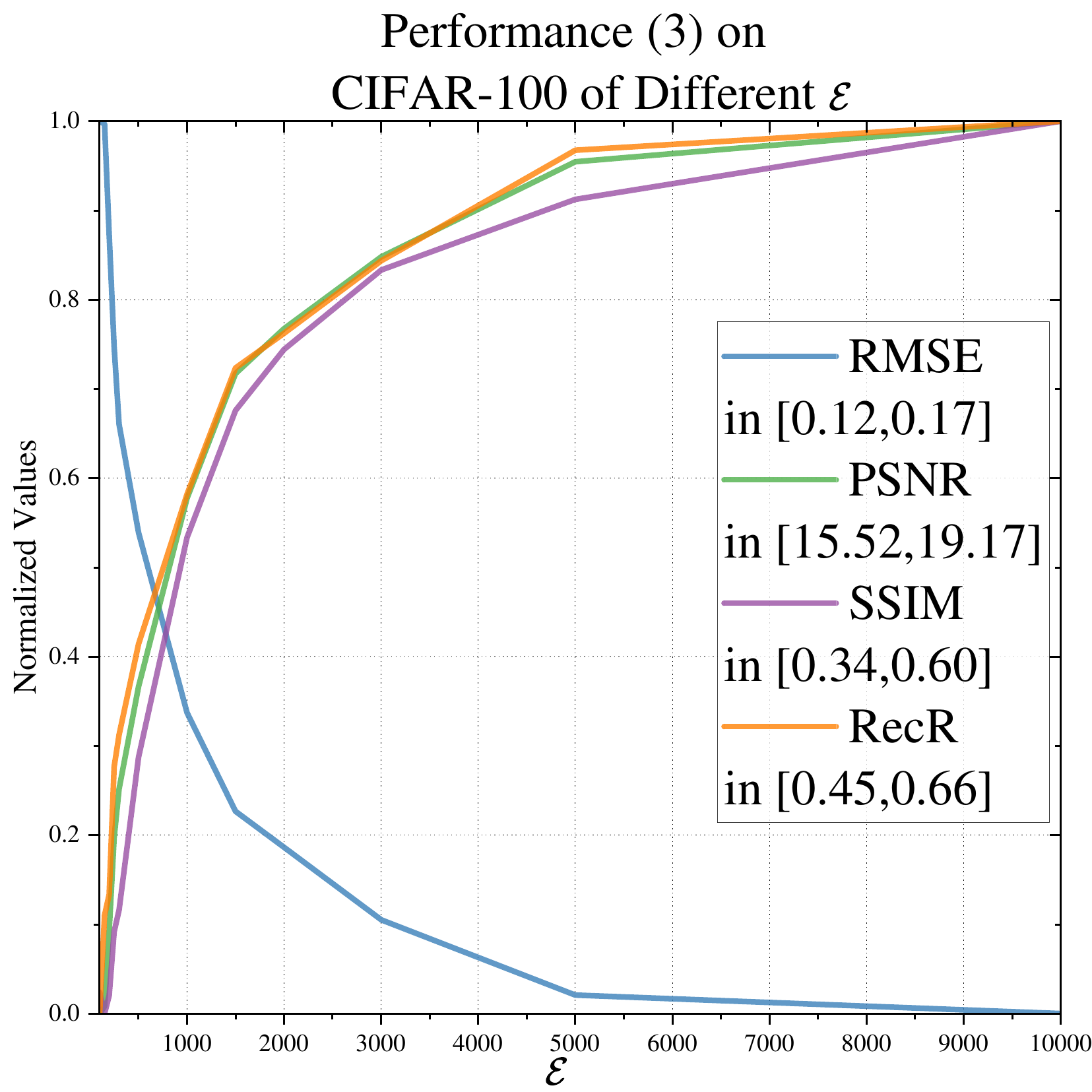}%
		\label{fig:E+_c100}}
	\hfil
	\subfloat[]{\includegraphics[width=0.13\textwidth]{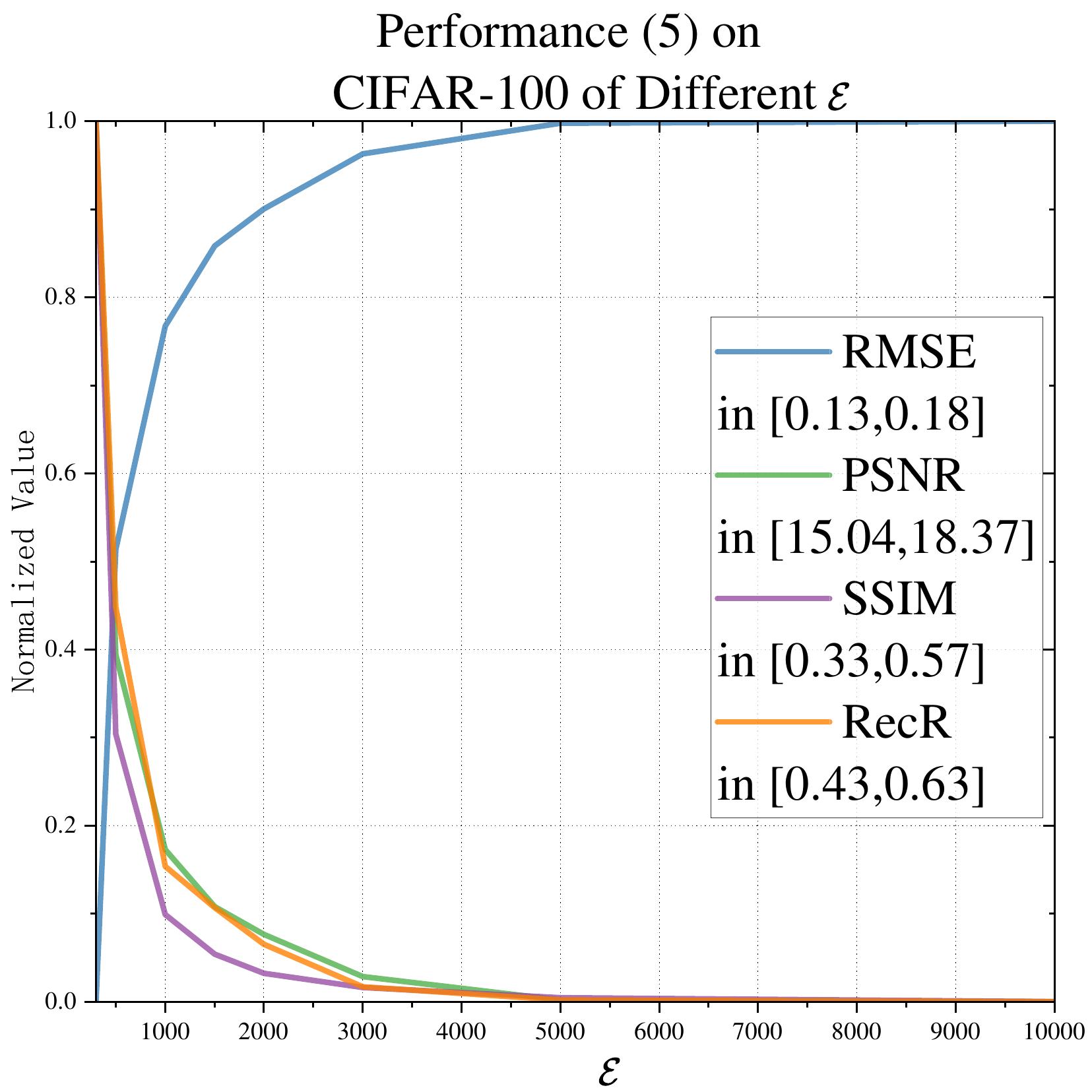}%
		\label{fig:E-_c100}}
	\caption{MAGIA performance in different Settings. The values are normalized based on maximum and minimum. (a), (b) and, (c) show $\mathcal{S}_{(1)}$, $\mathcal{S}_{(3)}$ and $\mathcal{S}_{(5)}$ on FEMNIST. (d), (e) and, (f) show $\mathcal{S}_{(1)}$, $\mathcal{S}_{(3)}$ and $\mathcal{S}_{(5)}$ on CIFAR-100.}
	\label{fig:ds}
\end{figure}


\begin{table*}[t!]
	\centering
	\setlength{\tabcolsep}{1mm}
	\caption{Comparison of reconstruction metrics for Top-1 performance of different $\mathcal{S}$. All data is rounded to two decimal places.\label{tab:s}}
	\small
	\begin{tabular}{cc|cccc|cccc}
		\toprule
		\multicolumn{2}{c|}{$|\mathcal{B}|=40$} & \multicolumn{4}{c|}{FEMNIST\_byclass} & \multicolumn{4}{c}{CIFAR-100} \\
		\cmidrule(r){3-6} \cmidrule(r){7-10}
		\multicolumn{2}{c|}{Top-1 $\mathcal{S}$}& $RMSE(\downarrow)$ & $PSNR(\uparrow)$ & $SSIM(\uparrow)$ & $RecR(\uparrow)$ & 
		$RMSE(\downarrow)$ & $PSNR(\uparrow)$ & $SSIM(\uparrow)$ & $RecR(\uparrow)$ \\
		\midrule
		\multicolumn{2}{c|}{MAGIA$_{(1)}$} & $\mathbf{0.13\pm0.06}$ & $\mathbf{19.17\pm4.76}$ & $\mathbf{0.69\pm0.20}$ & $\mathbf{0.79\pm0.12}$ & $\mathbf{0.11\pm0.04}$ & $\mathbf{19.67\pm2.95}$ & $\mathbf{0.63\pm0.12}$ & $\mathbf{0.68\pm0.17}$ \\
		\multicolumn{2}{c|}{MAGIA$_{(2)}$} & $0.21\pm0.03$ & $13.60\pm1.48$ & $0.39\pm0.12$ & $0.59\pm0.17$ & $0.16\pm0.04$ & $16.44\pm2.08$ & $0.37\pm0.10$ & $0.52\pm0.14$ \\
		\multicolumn{2}{c|}{MAGIA$_{(3)}$} & $0.13\pm0.06$ & $18.40\pm4.05$ & $0.68\pm0.19$ & $0.77\pm0.11$ & $0.12\pm0.04$ & $19.17\pm2.97$ & $0.60\pm0.12$ & $0.66\pm0.18$ \\
		\multicolumn{2}{c|}{MAGIA$_{(4,5)}$}  & $0.15\pm0.06$ & $17.35\pm4.32$ & $0.61\pm0.24$ & $0.69\pm0.16$ & $0.13\pm0.05$ & $18.37\pm3.07$ & $0.57\pm0.13$ & $0.63\pm0.17$ \\
		\bottomrule
	\end{tabular}
\end{table*}

\begin{table*}[t!]
	\centering
	\setlength{\tabcolsep}{1mm}
	\caption{Ablation Test of Reconstruction Metrics for Different Datasets. All data is rounded to two decimal places.\label{tab:at}}
	\small
	\begin{tabular}{cc|cccc|cccc}
		\toprule
		\multicolumn{2}{c|}{$|\mathcal{B}|=40$} & \multicolumn{4}{c|}{FEMNIST\_byclass} & \multicolumn{4}{c}{CIFAR-100} \\
		\cmidrule(r){3-6} \cmidrule(r){7-10}
		\multicolumn{2}{c|}{Ablations}&  $RMSE(\downarrow)$ & $PSNR(\uparrow)$ & $SSIM(\uparrow)$ & $RecR(\uparrow)$ & 
		$RMSE(\downarrow)$ & $PSNR(\uparrow)$ & $SSIM(\uparrow)$ & $RecR(\uparrow)$ \\
		\midrule
		\multicolumn{2}{c|}{MAGIA-a-m} & $0.17\pm0.07$ & $16.03\pm4.16$ & $0.57\pm0.21$ & $0.58\pm0.18$ & $0.29\pm0.04$ & $10.95\pm1.22$ & $0.25\pm0.07$ & $0.29\pm0.05$ \\
		\multicolumn{2}{c|}{MAGIA-a} & $0.19\pm0.07$ & $14.96\pm3.40$ & $0.54\pm0.19$ & $0.52\pm0.16$ & $0.30\pm0.04$ & $10.67\pm1.29$ & $0.27\pm0.07$ & $0.29\pm0.05$ \\
		\multicolumn{2}{c|}{MAGIA-m} & $0.13\pm0.06$ & $18.80\pm4.82$ & $0.67\pm0.21$ & $0.77\pm0.14$ & $0.12\pm0.04$ & $18.69\pm2.56$ & $0.54\pm0.12$ & $0.64\pm0.16$ \\
		\multicolumn{2}{c|}{\textbf{MAGIA}} & $\mathbf{0.13\pm0.06}$ & $\mathbf{19.17\pm4.76}$ & $\mathbf{0.69\pm0.20}$ & $\mathbf{0.79\pm0.12}$ & $\mathbf{0.11\pm0.04}$ & $\mathbf{19.67\pm2.95}$ & $\mathbf{0.63\pm0.12}$ & $\mathbf{0.68\pm0.17}$ \\
		\bottomrule
	\end{tabular}
\end{table*}


As shown in Fig.\ref{fig:ds}, with the exception of (d), MAGIA's $\mathcal{S}$ exhibits strong consistency and monotonicity across different methods. This behavior indicates that our optimization objective is well-designed, with a unique optimal solution that converges effectively. To further illustrate this, we have marked the value ranges of each metric in Fig.\ref{fig:ds}, enabling a clear observation of how the selection of $\mathcal{S}$ impacts performance. For (b) and (e), we observe that the growth of $\mathcal{E}$ has minimal impact on the final results, as its performance is comparable to $\mathcal{S}=\{1,2\}$. Similarly, in (c) and (f), the increase in $\mathcal{E}$ also has a reduced influence on the final results, with performance remaining close to $\mathcal{S}=\{\mathcal{B}-1,\mathcal{B}\}$. These trends align with the observations from (a) and (b), further validating the uniqueness and stability of the optimal solution in our approach.

Additionally, we present the Top-1 scores for each $\mathcal{S}$ selection method in Tab.\ref{tab:s}. Our analysis reveals that $\mathcal{S}$ exhibits dynamic behavior with iteration progression. It is worth noting that although the $PSNR$ of MAGIA$_{(1)}$ and MAGIA$_{(3)}$ is still a little far, the $SSIM$ and $RecR$ are actually very close. We supposed that this shows that MAGIA can be updated to an optimal value in a variety of ways. Different functional implementations of $\mathcal{S}$ updates can be tailored to particular use cases, lending strong generality and adaptability to our approach. These findings not only reaffirm the robustness of MAGIA's optimization objective but also underline its versatility in diverse scenarios. 




\subsubsection{Ablation Test}

The main contributions of our proposed MAGIA is the adaptive coefficient and momentum-based penalized term. To prove that our method was effective, we conducted an ablation test as Tab.\ref{tab:at}. In Tab.\ref{tab:at}, MAGIA-a-m means MAGIA without adaptive and momentum-based correction; MAGIA-a means MAGIA without adaptation; MAGIA-m means MAGIA without momentum-based penalized term.

As shown in Tab.\ref{tab:at}, our experiments reveal that the adaptive coefficient contributes more significantly to performance improvement compared to the momentum term. However, relying solely on the penalty term without incorporating the momentum term fails to achieve optimal results. This observation underscores the complementary roles of these components in MAGIA's design.

The adaptive coefficient plays a critical role in leveraging local optimal solutions to guide the optimization process, effectively aligning the global optimal solution within a more precise constraint boundary. Meanwhile, the momentum term enhances this process by stabilizing updates and improving convergence, particularly in challenging scenarios with larger batch sizes or complex gradient structures. Together, these components enable MAGIA to impose tighter constraints on gradient inversion, thereby substantially improving the effectiveness of GIAs.

%% file: appendixC.tex
\section{Conclusion and Futher Discussion}\label{appendix:c}

In this work we presented MAGIA, a conceptual framework motivated by our original goal of per–image gradient recovery under SAG and the practical constraint that only whole–batch averages are available. By invoking the triangle inequality over combinatorial partial–batch subsets and introducing a momentum–based correction together with an adaptive coefficient mechanism, we bridge the gap between the unattainable single–image inversion and the naive batch–level objective. This formulation lays the theoretical groundwork for probing how intra–batch gradient characteristics influence the global aggregated update, shifting focus from pure reconstruction accuracy to the fundamental information dynamics of gradient aggregation.

Despite progress in Gradient Inversion Attacks, three critical challenges remain: (1) overly large batch sizes still undermine attack efficacy—although MAGIA extends the feasible range, further novel gradient decomposition strategies inspired by our momentum and combinatorial constructs are needed; (2) the random client–side sampling in FedAvg exacerbates reconstruction variance—future work will integrate MAGIA’s partial–batch framework into asynchronous and heterogeneous settings, leveraging our adaptive coefficient to stabilize reconstructions across varying batch compositions; and (3) most studies rely on low–resolution datasets and simple architectures—bridging theory and practice demands applying MAGIA to high–resolution benchmarks and modern networks, as well as exploring hybrid defenses (e.g.\ combining MAGIA insights with secure aggregation or differential privacy) to evaluate robustness. We believe that deepening our understanding of how specific batch compositions correlate with information leakage will be crucial for designing federated learning systems that are both performant and inherently private.

%% file: arixv_main.bbl
\begin{thebibliography}{10}
\providecommand{\url}[1]{#1}
\csname url@samestyle\endcsname
\providecommand{\newblock}{\relax}
\providecommand{\bibinfo}[2]{#2}
\providecommand{\BIBentrySTDinterwordspacing}{\spaceskip=0pt\relax}
\providecommand{\BIBentryALTinterwordstretchfactor}{4}
\providecommand{\BIBentryALTinterwordspacing}{\spaceskip=\fontdimen2\font plus
\BIBentryALTinterwordstretchfactor\fontdimen3\font minus \fontdimen4\font\relax}
\providecommand{\BIBforeignlanguage}[2]{{%
\expandafter\ifx\csname l@#1\endcsname\relax
\typeout{** WARNING: IEEEtran.bst: No hyphenation pattern has been}%
\typeout{** loaded for the language `#1'. Using the pattern for}%
\typeout{** the default language instead.}%
\else
\language=\csname l@#1\endcsname
\fi
#2}}
\providecommand{\BIBdecl}{\relax}
\BIBdecl

\bibitem{mcmahan2017fedavg}
B.~McMahan, E.~Moore, D.~Ramage, S.~Hampson, and B.~A. y~Arcas, ``Communication-efficient learning of deep networks from decentralized data,'' in \emph{Artificial intelligence and statistics}.\hskip 1em plus 0.5em minus 0.4em\relax PMLR, 2017, pp. 1273--1282.

\bibitem{zhu2019dlg}
L.~Zhu, Z.~Liu, and S.~Han, ``Deep leakage from gradients,'' \emph{Advances in neural information processing systems}, vol.~32, 2019.

\bibitem{geiping2020inverting}
J.~Geiping, H.~Bauermeister, H.~Dr{\"o}ge, and M.~Moeller, ``Inverting gradients-how easy is it to break privacy in federated learning?'' \emph{Advances in neural information processing systems}, vol.~33, pp. 16\,937--16\,947, 2020.

\bibitem{yin2021gradinversion}
H.~Yin, A.~Mallya, A.~Vahdat, J.~M. Alvarez, J.~Kautz, and P.~Molchanov, ``See through gradients: Image batch recovery via gradinversion,'' in \emph{Proceedings of the IEEE/CVF conference on computer vision and pattern recognition}, 2021, pp. 16\,337--16\,346.

\bibitem{dimitrov2024spear}
D.~I. Dimitrov, M.~Baader, M.~M{\"u}ller, and M.~Vechev, ``Spear: Exact gradient inversion of batches in federated learning,'' \emph{Advances in Neural Information Processing Systems}, vol.~37, pp. 106\,768--106\,799, 2024.

\bibitem{carletti2025sok}
V.~Carletti, P.~Foggia, C.~Mazzocca, G.~Parrella, and M.~Vento, ``Sok: Gradient inversion attacks in federated learning.''

\bibitem{zhao2020idlg}
B.~Zhao, K.~R. Mopuri, and H.~Bilen, ``idlg: Improved deep leakage from gradients,'' \emph{arXiv preprint arXiv:2001.02610}, 2020.

\bibitem{yu2025gi}
W.~Yu, H.~Fang, B.~Chen, X.~Sui, C.~Chen, H.~Wu, S.-T. Xia, and K.~Xu, ``Gi-nas: Boosting gradient inversion attacks through adaptive neural architecture search,'' \emph{IEEE Transactions on Information Forensics and Security}, 2025.

\bibitem{tv}
L.~I. Rudin, S.~Osher, and E.~Fatemi, ``Nonlinear total variation based noise removal algorithms,'' \emph{Physica D: nonlinear phenomena}, vol.~60, no. 1-4, pp. 259--268, 1992.

\bibitem{leite2024federated}
L.~Leite, Y.~Santo, B.~L. Dalmazo, and A.~Riker, ``Federated learning under attack: Improving gradient inversion for batch of images,'' \emph{arXiv preprint arXiv:2409.17767}, 2024.

\bibitem{wang2024towards}
Y.~Wang, J.~Liang, and R.~He, ``Towards eliminating hard label constraints in gradient inversion attacks,'' \emph{arXiv preprint arXiv:2402.03124}, 2024.

\bibitem{caldas2018leaf}
S.~Caldas, S.~M.~K. Duddu, P.~Wu, T.~Li, J.~Kone{\v{c}}n{\`y}, H.~B. McMahan, V.~Smith, and A.~Talwalkar, ``Leaf: A benchmark for federated settings,'' \emph{arXiv preprint arXiv:1812.01097}, 2018.

\bibitem{cohen2017emnist}
G.~Cohen, S.~Afshar, J.~Tapson, and A.~Van~Schaik, ``Emnist: Extending mnist to handwritten letters,'' in \emph{2017 international joint conference on neural networks (IJCNN)}.\hskip 1em plus 0.5em minus 0.4em\relax IEEE, 2017, pp. 2921--2926.

\bibitem{netzer2011reading}
Y.~Netzer, T.~Wang, A.~Coates, A.~Bissacco, B.~Wu, A.~Y. Ng \emph{et~al.}, ``Reading digits in natural images with unsupervised feature learning,'' in \emph{NIPS workshop on deep learning and unsupervised feature learning}, vol. 2011, no.~5.\hskip 1em plus 0.5em minus 0.4em\relax Granada, 2011, p.~7.

\bibitem{krizhevsky2009learning}
A.~Krizhevsky, G.~Hinton \emph{et~al.}, ``Learning multiple layers of features from tiny images,'' 2009.

\bibitem{dimitrov2022dlf}
D.~I. Dimitrov, M.~Balunovic, N.~Konstantinov, and M.~Vechev, ``Data leakage in federated averaging,'' \emph{Transactions on Machine Learning Research}, 2022.

\bibitem{lre1}
Y.~Huang, S.~Gupta, Z.~Song, K.~Li, and S.~Arora, ``Evaluating gradient inversion attacks and defenses in federated learning,'' \emph{Advances in neural information processing systems}, vol.~34, pp. 7232--7241, 2021.

\bibitem{lti}
\BIBentryALTinterwordspacing
R.~Wu, X.~Chen, C.~Guo, and K.~Q. Weinberger, ``Learning to invert: Simple adaptive attacks for gradient inversion in federated learning,'' 2023. [Online]. Available: \url{https://arxiv.org/abs/2210.10880}
\BIBentrySTDinterwordspacing

\bibitem{dlg}
L.~Zhu, Z.~Liu, and S.~Han, ``Deep leakage from gradients,'' \emph{Advances in neural information processing systems}, vol.~32, 2019.

\bibitem{psnr}
A.~Hore and D.~Ziou, ``Image quality metrics: Psnr vs. ssim,'' in \emph{2010 20th international conference on pattern recognition}.\hskip 1em plus 0.5em minus 0.4em\relax IEEE, 2010, pp. 2366--2369.

\bibitem{dlf}
D.~I. Dimitrov, M.~Balunovic, N.~Konstantinov, and M.~Vechev, ``Data leakage in federated averaging,'' \emph{Transactions on Machine Learning Research}, 2022.

\bibitem{ig}
J.~Geiping, H.~Bauermeister, H.~Dr{\"o}ge, and M.~Moeller, ``Inverting gradients-how easy is it to break privacy in federated learning?'' \emph{Advances in neural information processing systems}, vol.~33, pp. 16\,937--16\,947, 2020.

\end{thebibliography}
